# Holistic, Instance-level Human Parsing

Qizhu Li*
qizhu.li@eng.ox.ac.uk

Anurag Arnab*
anurag.arnab@eng.ox.ac.uk

Philip H.S. Torr
philip.torr@eng.ox.ac.uk

Department of Engineering Science
University of Oxford
Oxford, UK

## Abstract

Object parsing – the task of decomposing an object into its semantic parts – has traditionally been formulated as a category-level segmentation problem. Consequently, when there are multiple objects in an image, current methods cannot count the number of objects in the scene, nor can they determine which part belongs to which object. We address this problem by segmenting the parts of objects at an instance-level, such that each pixel in the image is assigned a part label, as well as the identity of the object it belongs to. Moreover, we show how this approach benefits us in obtaining segmentations at coarser granularities as well. Our proposed network is trained end-to-end given detections, and begins with a category-level segmentation module. Thereafter, a differentiable Conditional Random Field, defined over a variable number of instances for every input image, reasons about the identity of each part by associating it with a human detection. In contrast to other approaches, our method can handle the varying number of people in each image and our holistic network produces state-of-the-art results in instance-level part and human segmentation, together with competitive results in category-level part segmentation, all achieved by a single forward-pass through our neural network.

## 1 Introduction

Object parsing, the segmentation of an object into semantic parts, is naturally performed by humans to obtain a more detailed understanding of the scene. When performed automatically by computers, it has many practical applications, such as in human-robot interaction, human behaviour analysis and image descriptions for the visually impaired. Furthermore, detailed part information has been shown to be beneficial in other visual recognition tasks such as fine-grained recognition [47], human pose estimation [13] and object detection [37]. In this paper, we focus on the application of parsing humans as it is more commonly studied, although our method makes no assumptions on the type of object it is segmenting.

In contrast to existing human parsing approaches [18, 29, 45], we operate at an *instance level* (to our knowledge, we are the first work to do so). As shown in Fig. 1, not only do we segment the various body parts of humans (Fig. 1b), but we associate each of these parts to one of the humans in the scene (Fig. 1c), which is particularly important for understanding scenes with multiple people. In contrast to existing instance segmentation work [10, 31, 34], we operate at a more detailed part level, enabling us to extract more comprehensive information of the scene. Furthermore, with our part-level instance segmentation of humans, we can easily recover human-level instance segmentation (by taking the union of all parts assigned to a particular instance as shown in Fig. 1d), and we show significant improvement over previous state-of-the-art in human instance-segmentation when doing so.

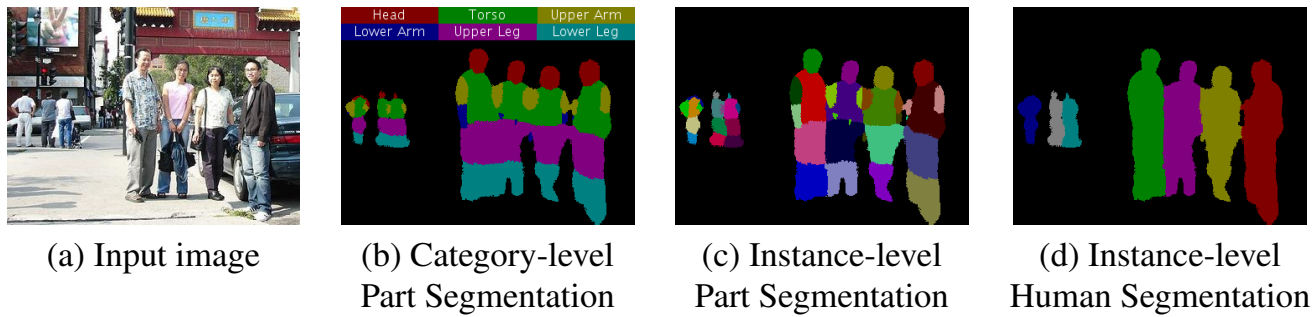

| (a) Input image | (b) Category-level Part Segmentation | (c) Instance-level Part Segmentation | (d) Instance-level Human Segmentation |

Figure 1: Our proposed approach segments human parts at an instance level (c) (which to our knowledge is the first work to do so) from category-level part segmentations produced earlier in the network (b). Moreover, we can easily obtain human instance segmentations (d) by taking the union of all pixels associated to a particular person. Therefore, our proposed end-to-end trained neural network parses humans into semantic parts at both category and instance level in a single forward-pass. Best viewed in colour.

Our approach is based on a deep Convolutional Neural Network (CNN), which consists of an initial category-level part segmentation module. Using the output of a human detector, we are then able to associate segmented parts with detected humans in the image using a differentiable Conditional Random Field (CRF), producing a part-level instance segmentation of the image. Our formulation is robust to false-positive detections as well as imperfect bounding boxes which do not cover the entire human, in contrast to other instance segmentation methods based on object detectors [10, 20, 21, 26, 34]. Given object detections, our network is trained end-to-end, given detections, with a novel loss function which allows us to handle a variable number of human instances on every image.

We evaluate our approach on the Pascal Person-Parts [8] dataset, which contains humans in a diverse set of poses and occlusions. We achieve state-of-the-art results on instance-level segmentation of both body parts and humans. Moreover, our results on semantic part segmentation (which is not-instance aware) is also competitive with current state-of-the-art. All of these results are achieved with a holistic, end-to-end trained model which parses humans at both an instance and category level, and outputs a dynamic number of instances per image, all in a single forward-pass through the network.

## 2   Related Work

The problem of object parsing, which aims to decompose objects into their semantic parts, has been addressed by numerous works [27, 29, 38, 43, 45], most of which have concentrated on parsing humans. However, none of the aforementioned works have parsed objects at an instance level as shown in Fig. 1, but rather category level. In fact, a lot of work on human parsing has focussed on datasets such as Fashionista [46], ATR [27] and Deep Fashion [35] where images typically contain only one, centred person. The notion of instance-level segmentation only matters when more than one person is present in an image, motivating us to evaluate our method on the Pascal Person-Parts dataset [8] where multiple people can appear in unconstrained environments. Recent human parsing approaches have typically been similar to semantic segmentation works using fully convolutional networks (FCNs) [36], but trained to label parts [5, 6, 7] instead of object classes. However, methods using only FCNs do not explicitly model the structure of a human body, and typically do not perform as well as methods which do [29]. Structural priors of the human body have been encoded using pictorial structures [15, 17], Conditional Random Fields (CRFs) [4, 23, 25, 43] and more recently, with LSTMs [29, 30]. The HAZN approach of [45] addressed the problem that some parts are often very small compared to other parts and difficult to segment with scale-variant CNNs. This scale variation was handled by a cascade of three separately-trained FCNs, each parsing different regions of the image at different scales.

An early instance segmentation work by Winn *et al.* [44] predicted the parts of an object, and then encouraged these parts to maintain a spatial ordering, characteristic of an instance, using asymmetric pairwise potentials in a CRF. However, subsequent work has not operated at a part level. Zhang *et al.* [48, 49] performed instance segmentation of vehicles using an MRF. However, this graphical model was not trained end-to-end as done by [3, 32, 51] and our approach. Furthermore, they assumed a maximum of 9 cars per image. Approaches using recurrent neural networks [39, 40] can handle a variable number of instances per image by segmenting an instance per time-step, but are currently restricted to only one object category. Our method, on the other hand, is able to handle both an arbitrary number of objects, and multiple object categories in the image with a single forward-pass through the network.

Various methods of instance segmentation have also involved modifying object detection systems to output segments instead of bounding boxes [10, 20, 21, 26]. However, these methods cannot produce a segmentation map of the image, as shown in Fig. 1, without post-processing as they consider each detection independently. Although our method also uses an object detector, it considers all detections in the image jointly with an initial category-level segmentation, and produces segmentation maps naturally where one pixel cannot belong to multiple instances in contrast to the aforementioned approaches. The idea of combining the outputs of a category-level segmentation network and an object detector to reason about different instances was also presented by [1]. However, that system was not trained end-to-end, could not segment instances outside the detector's bounding box, and did not operate at a part level.

# 3 Proposed Approach

Our network (Fig. 2) consists of two components: a category-level part segmentation module, and an instance segmentation module. As both of these modules are differentiable, they can be integrated into a single network and trained jointly. The instance segmentation module (Sec. 3.2) uses the output of the first category-level segmentation module (Sec. 3.1) as well as the outputs of an object detector as its input. It associates each pixel in the category-level segmentation with an object detection, resulting in an instance-level segmentation of the image. Given a $H \times W \times 3$ input image, $\mathbf{I}$, the category-level part segmentation module produces a $H \times W \times (P+1)$ dimensional output $\mathbf{Q}$ where $P$ is the number of part classes in the dataset and one background class. There can be a variable number, $D$, of human detections per image, and the output of the instance segmentation module is an $H \times W \times (PD+1)$ tensor denoting the probabilities, at each pixel in the image, of each of the $P$ part classes belonging to one of the $D$ detections.

Two challenges of instance segmentation are the variable number of instances in every image, and the fact that permutations of instance labels lead to identical results (in Fig. 1, how we order the different people does not matter). Zhang *et al.* [48, 49] resolve these issues by assuming a maximum number of instances and using the ground-truth depth ordering of instances respectively. Others have bypassed both of these issues by predicting each instance independently [10, 20, 21, 26], but this also allows a pixel to belong to multiple instances. Instead, we use a loss function (Sec 3.3) that is based on "matching" the prediction to the ground-truth, allowing us to handle permutations of the ground truth. Furthermore, weight-sharing in our instance segmentation module allows us to segment a variable number of instances per image. As a result, we do not assume a maximum number of instances, consider all instances jointly, and train our network end-to-end, given object detections.

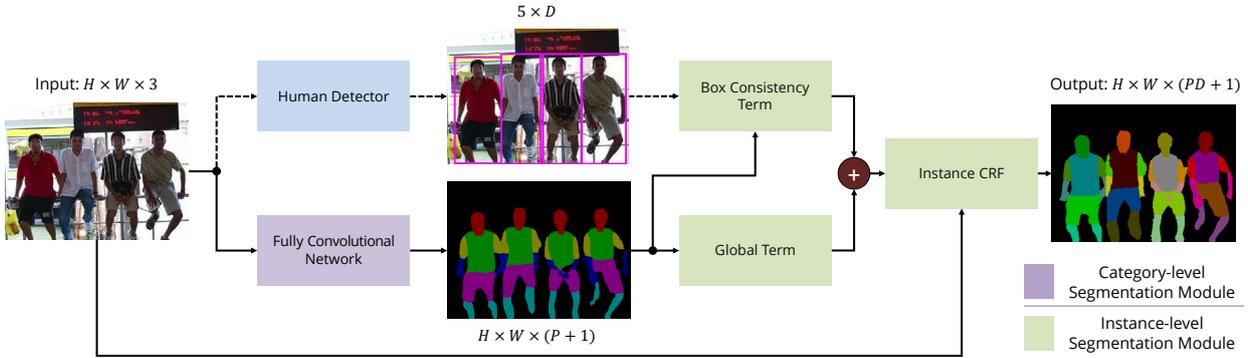

Figure 2: Our proposed approach. An $H \times W \times 3$ image is input to a human detection network and a body parts semantic segmentation network, producing $D$ detections of human and an $H \times W \times (P+1)$ dimensional feature map respectively, where $(P+1)$ is the size of the semantic label space including a background class. These results are used to form the unary potentials of an Instance CRF which performs instance segmentation by associating labelled pixels with human detections. In the above diagram, dotted lines represent forward only paths, and solid lines show routes where both features and gradients flow. The green boxes form the instance-level segmentation module (Sec. 3.2). Best viewed in colour.

## 3.1 Category-level part segmentation module

The part segmentation module is a fully convolutional network [36] based on ResNet-101 [22]. A common technique, presented in [6, 7], is to predict the image at three different scales (with network weights shared among all the scales), and combine predictions together with learned, image-dependent weights. We take a different approach of fusing information at multiple scales – we pool the features after `res5c` [22] at five different resolutions (by varying the pooling stride), upsample the features to the resolution before pooling, and then concatenate these features before passing them to the final convolutional classifier, as proposed in [50]. As we show in Sec 4.4, this approach achieves better semantic segmentation results than [6, 7]. We denote the output of this module by the tensor, $\mathbf{Q}$, where $Q_i(l)$ is the probability of pixel $i$ being assigned label $l \in \{0, 1, 2, ..., P\}$. Further details of this module are included in the appendix.

## 3.2 Instance-level segmentation module

This module creates an instance-level segmentation of the image by associating each pixel in the input category-level segmentation, $\mathbf{Q}$, with one of the $D$ input human-detections or the background label. Let there be $D$ input human-detections for the image, where the $i$-th detection is represented by $B_i$, the set of pixels lying within the four corners of its bounding box, and $s_i \in [0, 1]$, the detection score. We assume that the 0-th detection refers to the background label. Furthermore, we define a multinomial random variable, $V_i$, at each of the $N$ pixels in the image, and let $\mathbf{V} = [V_1, V_2, ..., V_N]^\top$. This variable can take on a label from the set $\{1, 2, ..., D\} \times \{1, 2, ..., P\} \cup \{(0,0)\}$ since each of the $P$ part labels can be associated with one of the $D$ human detections, or that pixel could belong to the background label, $(0,0)$.

We formulate a Conditional Random Field over these $V$ variables, where the energy of the assignment $\mathbf{v}$ to all of the instance variables $\mathbf{V}$ consists of two unary terms, and one pairwise term (whose weighting co-efficients are all learned via backpropagation):

$$E(\mathbf{V} = \mathbf{v}) = -\sum_{i}^{N} \ln \left[ w_1 \psi_{Box}(v_i) + w_2 \psi_{Global}(v_i) + \varepsilon \right] + \sum_{i<j}^{N} \psi_{Pairwise}(v_i, v_j). \quad (1)$$

The unary and pairwise potentials are computed within our neural network, differentiable with respect to their input and parameters, and described in Sec. 3.2.1 through 3.2.3. The Maximum-a-Posteriori (MAP) estimate of our CRF (since the energy in Eq. 1 characterises a Gibbs distribution) is computed as the final labelling produced by our network. We perform the iterative mean-field inference algorithm to approximately compute the MAP solution by minimising Eq. 1. As shown by Zheng *et al.* [51], this can be formulated as a Recurrent Neural Network (RNN), allowing it to be trained end-to-end as part of a larger network. However, as our network is input a variable number of detections per image, $D$, the label space of the CRF is dynamic. Therefore, unlike [51], the parameters of our CRF are not class-specific to allow for this variable number of "channels".

### 3.2.1 Box Consistency Term

We observe that in most cases, a body part belonging to a person is located inside the bounding box of the person. Based on this observation, the box consistency term is employed to encourage pixel locations inside a human bounding box $B_i$ to be associated with the $i$-th human detection. The box term potential at spatial location $k$ for body part $j$ of a human $i$ is assigned either 0 for $k \notin B_i$, or the product of the detection score, $s_i$, and the category-level part segmentation confidence, $Q_k(j)$, for $k \in B_i$. For $(i, j) \in \{1, 2, ..., D\} \times \{1, 2, ..., P\}$,

$$\psi_{Box}(V_k = (i, j)) = \begin{cases} s_i Q_k(j) & \text{if } k \in B_i \\ 0 & \text{otherwise.} \end{cases} \quad (2)$$

Note that this potential may be robust to false-positive detections when the category-level segmentation and human detection do not agree with each other, since $Q_k(l)$, the probability of a pixel $k$ taking on body-part label $l$, is low. Furthermore, note that we use one human-detection to reason about the identity of all parts which constitute that human.

### 3.2.2 Global Term

A possible shortcoming for the box consistency potential is that if some pixels belonging to a human instance fall outside the bounding box and are consequently assigned 0 for the box consistency term potential, they would be lost in the final instance segmentation prediction. Visually, the generated instance masks would appear truncated along the bounding box boundaries – a problem suffered by [1, 10, 21, 26]. To overcome this undesirable effect, we introduce the global potential: it complements the box consistency term by assuming that a pixel is equally likely to belong to any one of the detected humans. It is expressed as

$$\psi_{Global}(V_k = (i, j)) = Q_k(j), \quad (3)$$

for $(i, j) \in \{1, 2, ..., D\} \times \{1, 2, ..., P\} \cup \{(0,0)\}$.

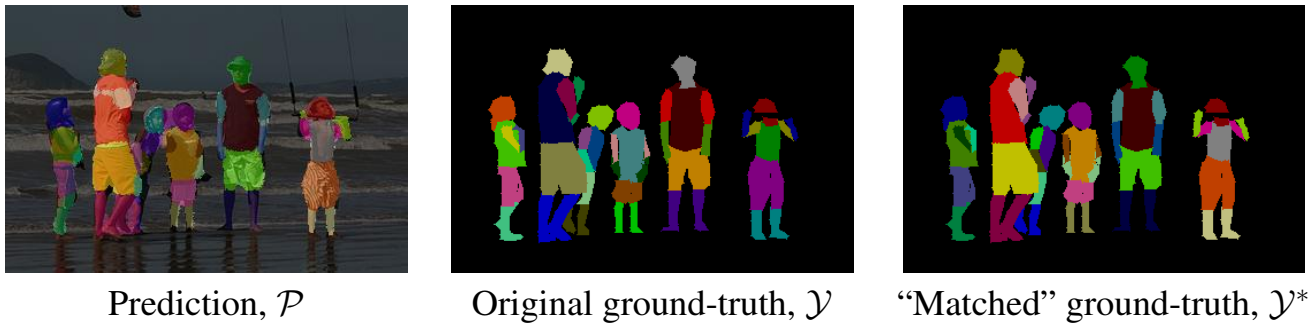

Prediction, $\mathcal{P}$          Original ground-truth, $\mathcal{Y}$          "Matched" ground-truth, $\mathcal{Y}^*$

Figure 3: As different permutations of the ground-truth are equivalent in the case of instance segmentation, we "match" the original ground-truth, $\mathcal{Y}$, to our network's prediction, $\mathcal{P}$, to obtain the "matched" ground-truth which we use to compute our loss during training.

### 3.2.3 Pairwise Term

Our pairwise term is composed of densely-connected Gaussian kernels [24] which are commonly used in segmentation literature [5, 51]. This pairwise potential encourages both spatial and appearance consistency, and we find these priors to be suitable in the case of instance-level segmentation as well. As in [51], the weighting parameters of these potentials are learned via backpropagation, though in our case, the weights are shared among all classes.

## 3.3 Loss function and network training

We first pre-train the category-level segmentation part of our network, as described in the appendix. Thereafter, we add the instance segmentation module, and train with a permutation-invariant loss function which is backpropagated through both our instance- and category-level segmentation networks. Since all permutations of an instance segmentation have the same qualitative result, we "match" the original ground-truth to our prediction before computing the loss, as shown in Fig. 3. This matching is based on the Intersection over Union (IoU) [14] of a predicted and ground-truth instance, similar to [40]. Let $\mathcal{Y} = \{y_1, y_2, ..., y_m\}$, a set of $m$ segments, denote the ground-truth labelling of an image, where each segment is an instance and has a part label assigned to it. Similarly, let $\mathcal{P} = \{p_1, p_2, ..., p_n\}$ denote our $n$ predicted instances, each with an associated part label. Note that $m$ and $n$ need not be the same as we may predict greater or fewer instances than there actually are in the image. The "matched" ground truth, $\mathcal{Y}^*$ is the permutation of the original ground-truth labelling which maximises the IoU between our prediction, $\mathcal{P}$ and ground-truth

$$\mathcal{Y}^* = \underset{\mathcal{Z} \in \pi(\mathcal{Y})}{\arg\max} \, \mathrm{IoU}(\mathcal{Z}, \mathcal{P}), \qquad (4)$$

where $\pi(\mathcal{Y})$ denotes the set of all permutations of $\mathcal{Y}$. Note that we define the IoU between all segments of different labels to be 0. Eq. 4 can be solved efficiently using the Hungarian algorithm as it can be formulated as a bipartite graph matching problem, and once we have the "matched" ground-truth, $\mathcal{Y}^*$, we can apply any loss function to it and train our network for segmentation.

In our case, we use the standard cross-entropy loss function on the "matched" ground truth. In addition, we employ Online Hard Example Mining (OHEM), and only compute our loss over the top $K$ pixels with the highest loss in the training mini-batch. We found that during training, many pixels already had a high probability of being assigned to the correct class. By only selecting the top $K$ pixels with the highest loss, we are able to encourage our network to improve on the pixels it is currently misclassifying, as opposed to increasing the probability of a pixel it is already classifying correctly. This approach was inspired by "bootstrapping" [12, 42] or "hard-negative mining" [16] commonly used in training object detectors. However, these methods mined hard examples from the entire dataset. Our approach is most similar to [41], who mined hard examples online from each mini-batch in the context of detection. Similar to the aforementioned works, we found OHEM to improve our overall results, as shown in Sec. 4.2.

## 3.4 Obtaining segmentations at other granularities

Given the part instance prediction produced by our proposed network, we are able to easily obtain human instance segmentation and semantic part segmentation. In order to achieve human instance segmentation, we map the predicted part instance labels $(i, j)$, i.e. part $j$ of person $i$, to $i$. Whereas to obtain semantic part segmentation, we map predicted part instance labels $(i, j)$ to $j$ instead.

# 4 Experiments

We describe our dataset and experimental set-up in Sec. 4.1, before presenting results on instance-level part segmentation (Fig. 1c), instance-level human segmentation (Fig. 1d) and semantic part segmentation (Fig. 1b). Additional quantitative and qualitative results, failure cases and experimental details are included in the appendix.

## 4.1 Experimental Set-up

We evaluate our proposed method on the Pascal Person-Part dataset [13] which contains 1716 training images, and 1817 test images. This dataset contains multiple people per image in unconstrained poses and environments, and contains six human body part classes (Fig. 1b), as well as the background label. As described in Sec. 3.3, we initially pre-train our category-level segmentation module before training for instance-level segmentation. This module is first trained on the 21 classes of the Pascal VOC dataset [14], and then finetuned on the seven classes of the Pascal Part training set using category-level annotations. Finally, we train for instance segmentation with instance-level ground truth. Full details of our training process, including all hyperparameters such as learning rate, are in the appendix. To clarify these details, we will also release our code.

We use the standard $AP^r$ metric [20] for evaluating instance-level segmentation: the mean Average Precision of our predictions is computed where a prediction is considered correct if its IoU with a ground-truth instance is above a certain threshold. This is similar to the $AP$ metric used in object detection. However, in detection, the IoU between ground-truth and predicted bounding boxes is computed, whereas here, the IoU between regions is computed. Furthermore, in detection, an overlap threshold of 0.5 is used, whereas we vary this threshold. Finally, we define the $AP^r_{vol}$ which is the mean of the $AP^r$ score for overlap thresholds varying from 0.1 to 0.9 in increments of 0.1.

We use the publicly available R-FCN detection framework [11], and train a new model with data from VOC 2012 [14] that do not overlap with any of our test sets. We train with all object classes of VOC, and only use the output for the human class. Non-maximal suppression is performed on all detections before being fed into our network.

Table 1: Comparison of $AP^r$ at various IoU thresholds for instance-level part segmentation on the Pascal Person-Parts dataset

| Method | IoU threshold | | | $AP^r_{vol}$ |
| | 0.5 | 0.6 | 0.7 | |
| --- | --- | --- | --- | --- |
| MNC [10] | 38.8 | 28.1 | **19.3** | 36.7 |
| Ours, piecewise trained, box term only* | 38.0 | 27.4 | 16.7 | 36.6 |
| Ours, piecewise trained | 38.8 | 28.5 | 17.6 | 37.3 |
| Ours, end-to-end trained | 39.0 | 28.6 | 17.4 | 37.7 |
| Ours, piecewise trained, box term only, OHEM | 38.7 | 28.9 | 17.5 | 36.7 |
| Ours, piecewise trained, OHEM | 39.7 | 29.7 | 18.7 | 37.4 |
| Ours, end-to-end trained, OHEM | **40.6** | **30.4** | 19.1 | **38.4** |

*Model is equivalent to our reimplementation of [1]

## 4.2 Results on Instance-level Part Segmentation

Table 1 shows our results on part-level instance segmentation on the Pascal Person-Part dataset. To our knowledge, we are the first work to do this, and hence we study the effects of various design choices on overall performance. We also use the publicly available code for MNC [10], which won the MS-COCO 2016 instance segmentation challenge, and finetune their public model trained on VOC 2011 [19] on Person-Part instances as a baseline.

We first train our model in a piecewise manner, by first optimising the parameters of the category-level segmentation module, and then "freezing" the weights of this module and only training the instance network. Initially, we only use the box consistency term (Sec. 3.2.1) in the Instance CRF, resulting in an $AP^r$ at 0.5 of 38.0%. Note that this model is equivalent to our reimplementation of [1]. Adding in the global potential (Sec. 3.2.2) helps us cope with bounding boxes which do not cover the whole human, and we see an improvement at all IoU thresholds. Training our entire network end-to-end gives further benefits. We then train all variants of our model with OHEM, and observe consistent improvements across all IoU thresholds with respect to the corresponding baseline. Here, we set $K = 2^{15}$, meaning that we computed our loss over $2^{15}$ or approximately 12% of the hardest pixels in each training image (since we train at full resolution). We also employ OHEM when pre-training the category-level segmentation module of our network, and observe minimal difference in the final result if we use OHEM when training the category-level segmentation module but not the instance segmentation module. Training end-to-end with OHEM achieves 2.6% higher in $AP^r$ at 0.5, and 1.8% higher $AP^r_{vol}$ over a piecewise-trained baseline model without OHEM and only the box term (second row), which is equivalent to the model of [1]. Furthermore, our $AP^r_{vol}$ is 1.7% greater than the strong MNC [10] baseline. Note that although [21] also performed instance-level segmentation on the same dataset, their evaluation was only done using human instance labels, which is similar to our following experiment on human instance segmentation.

## 4.3 Results on Human Instance Segmentation

We can trivially obtain instance-level segmentations of humans (Fig 1d), as mentioned in Sec. 3.4. Table 2 shows our state-of-the-art instance segmentation results for humans on the VOC 2012 validation set [14]. We use the best model from the previous section as there is no overlap between the Pascal Person-Part training set, and the VOC 2012 validation set.

Table 2: Comparison of $AP^r$ at various IoU thresholds for instance-level human segmentation on the VOC 2012 validation set

| Method | IoU threshold | | | | | $AP^r_{vol}$ |
| | 0.5 | 0.6 | 0.7 | 0.8 | 0.9 | |
|---|---|---|---|---|---|---|
| SDS [20] | 47.9 | 31.8 | 15.7 | 3.3 | 0.1 | – |
| Chen *et al.* [9] | 48.3 | 35.6 | 22.6 | 6.5 | 0.6 | – |
| PFN [28] | 48.4 | 38.0 | 26.5 | 16.5 | 5.9 | 41.3 |
| Arnab *et al.* [1]* | 58.6 | 52.6 | 41.1 | 30.4 | 10.7 | 51.8 |
| R2-IOS [31] | 60.4 | 51.2 | 33.2 | – | – | – |
| Arnab *et al.* [2]* | 65.6 | 58.0 | 46.7 | 33.0 | 14.6 | 57.4 |
| Ours, piecewise | 64.0 | 59.8 | 51.0 | 38.3 | **20.1** | 57.2 |
| Ours, end-to-end | **70.2** | **63.1** | **54.1** | **41.0** | 19.6 | **61.0** |

\*Results obtained from supplementary material.

Table 3: Comparison of semantic part segmentation results on the Pascal Person-Parts test set

| Method | IoU [%] |
|---|---|
| DeepLab* [5] | 53.0 |
| Attention [7] | 56.4 |
| HAZN [45] | 57.5 |
| LG-LSTM [30] | 58.0 |
| Graph LSTM [29] | 60.2 |
| DeepLab v2 [6] | 64.9 |
| RefineNet [33] | 68.6 |
| Ours, pre-trained | 65.9 |
| Ours, final network | 66.3 |

\*Result reported in [45]

As Tab. 2 shows, our proposed approach outperforms previous state-of-the-art by a significant margin, particularly at high IoU thresholds. Our model receives extra supervision in its part labels, but the fact that our network can implicitly infer relationships between different parts whilst training may help it handle occluding instances better than other approaches, leading to better instance segmentation performance. The fact that our network is trained with part-level annotations may also help it identify small features of humans better, leading to more precise segmentations and thus improvements at high $AP^r$ thresholds. Our $AP^r$ at each IoU threshold for human instance segmentation is higher than that for part instance segmentation (Tab. 1). This is because parts are smaller than entire humans, and thus more difficult to localise accurately. An alternate method of performing instance-level part segmentation may be to first obtain an instance-level human segmentation using another method from Tab. 2, and then partition it into the various body parts of a human. However, our approach, which groups parts into instances, is validated by the fact that it achieves state-of-the-art instance-level human segmentation performance.

### 4.4    Results on Category-level Part Segmentation

Finally, our model is also able to produce category-level segmentations (as shown in Fig. 1b). This can be obtained from the output of the category-level segmentation module, or from our instance module as described in Sec. 3.4. As shown in Tab. 3, our semantic segmentation results are competitive with current state-of-the-art. By training our entire network consisting of the category-level and instance-level segmentation modules jointly, and then obtaining the semantic segmentation from the final instance segmentation output by our network, we are able to obtain a small improvement of 0.4% in mean IoU over the output of the initial semantic segmentation module.

## 5    Conclusion

Our proposed, end-to-end trained network outputs instance-level body part and human segmentations, as well as category-level part segmentations in a single forward-pass. Moreover, we have shown how segmenting objects into their constituent parts helps us segment the object as a whole with our state-of-the-art results on instance-level segmentation of both body parts and entire humans. Furthermore, our category-level segmentations improve after training for instance-level segmentation. Our future work is to train the object detector end-to-end as well. Moreover, the improvement that we obtained in instance segmentation of humans as a result of first segmenting parts motivates us to explore weakly-supervised methods which do not require explicit object part annotations.

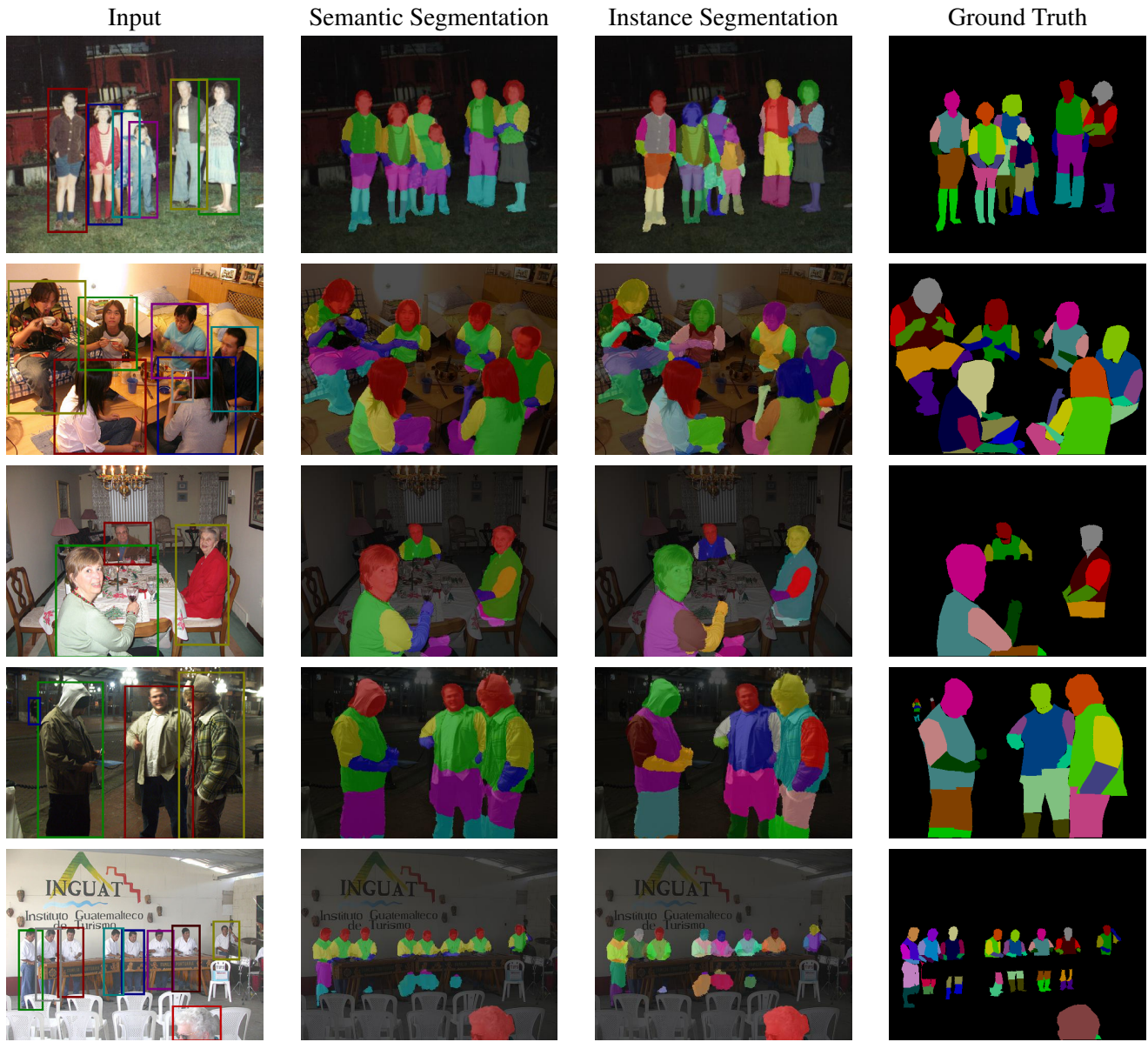

Figure 4: Some results of our system. The first column shows the input image and the input detections we obtained from training the R-FCN detector [11]. The second and third columns show our final semantic segmentation (Sec. 3.4) and instance-level part segmentation. *First row*: our network can deal with poor bounding box localisation, as it manages to segment the third person from the left although the bounding box only partially covers her. *Second row*: our method is robust against false positive detections because of the box term. Observe that the bowl of the rightmost person in the bottom row is falsely detected as a person, but rejected in the final prediction. *Following rows*: we are able to handle overlapping bounding boxes by reasoning globally using the Instance CRF.

**Acknowledgement**    We thank Stuart Golodetz for discussions and feedback. This work was supported by the EPSRC, Clarendon Fund, ERC grant ERC-2012-AdG 321162-HELIOS, EPSRC grant Seebibyte EP/M013774/1 and EPSRC/MURI grant EP/N019474/1.

# Appendix

In this appendix, we present additional results of our proposed approach in Sec. A, and provide additional training and implementation details in Sec. B (both for our model, and the strong MNC baseline [10]).

# A    Additional Results

In our main paper, we reported our $AP^r$ results averaged over all classes. Fig. 5 visualises the per-class results of our best model at different IoU thresholds. Fig. 6 displays the success cases of our method, while Fig. 7 shows examples of failure cases. Furthermore, we illustrate the strengths and weaknesses of our part instance segmentation method in comparison to MNC [10] in Fig. 8, and compare our instance-level human segmentation results, which we obtain by the simple mapping described in Sec. 3.4 of our main paper, to MNC in Fig. 9.

Finally, we attach an additional video. We run our system offline, on a frame-by-frame basis on the entire music video, and show how our method is able to accurately parse humans at both category and instance level on internet data outside the Pascal dataset. Instance-level segmentation of videos requires data association. We use a simple, greedy method which operates on a frame-by-frame basis. Segments from one frame are associated to segments in the next frame based on the IoU, using the same method we use for our loss function as described in Sec. 3.3 of the main paper.

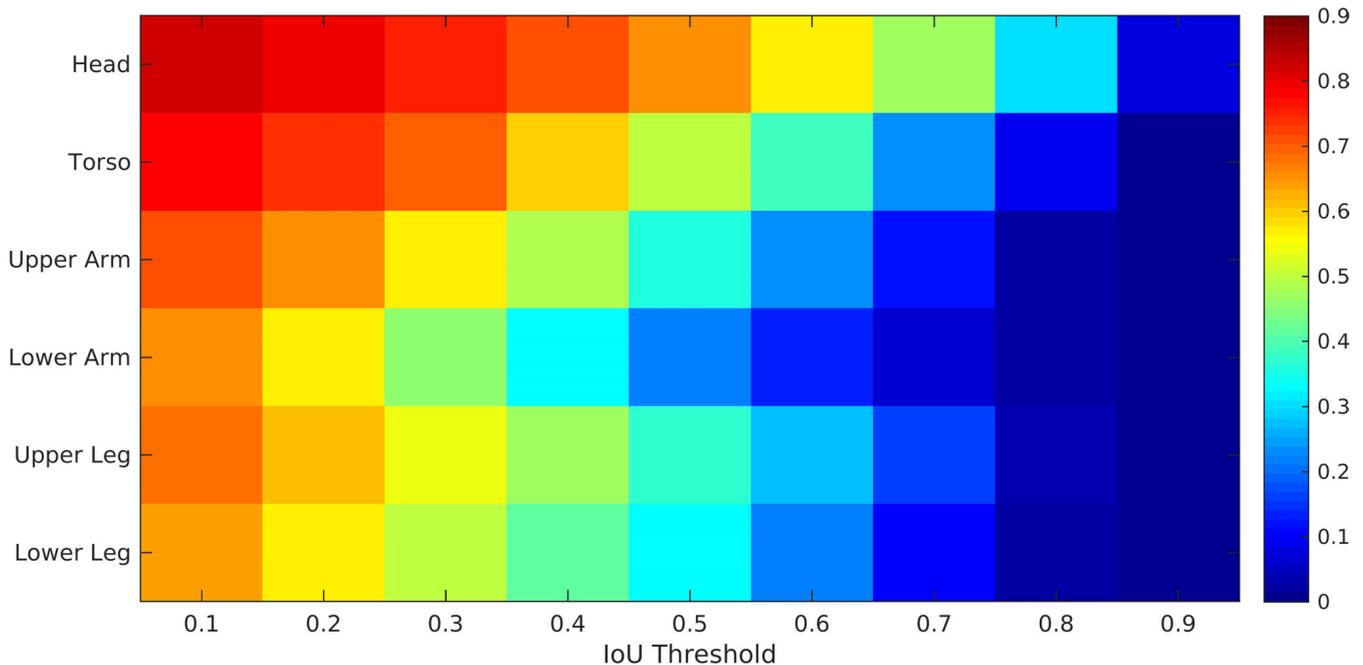

Figure 5: **Visualisation of per-class results for different IoU thresholds on the Pascal Person-Parts test set.** The heatmap shows the per-class $AP^r$ of our best model at IoU thresholds from 0.1 to 0.9 in increments of 0.1 on the Pascal Person-Parts test set. It shows that our method achieves best instance accuracy for the head category, and finds lower arms and lower legs most challenging to segment correctly. This is likely because of the thin shape of the lower limbs which is known to pose difficulty for semantic segmentation.

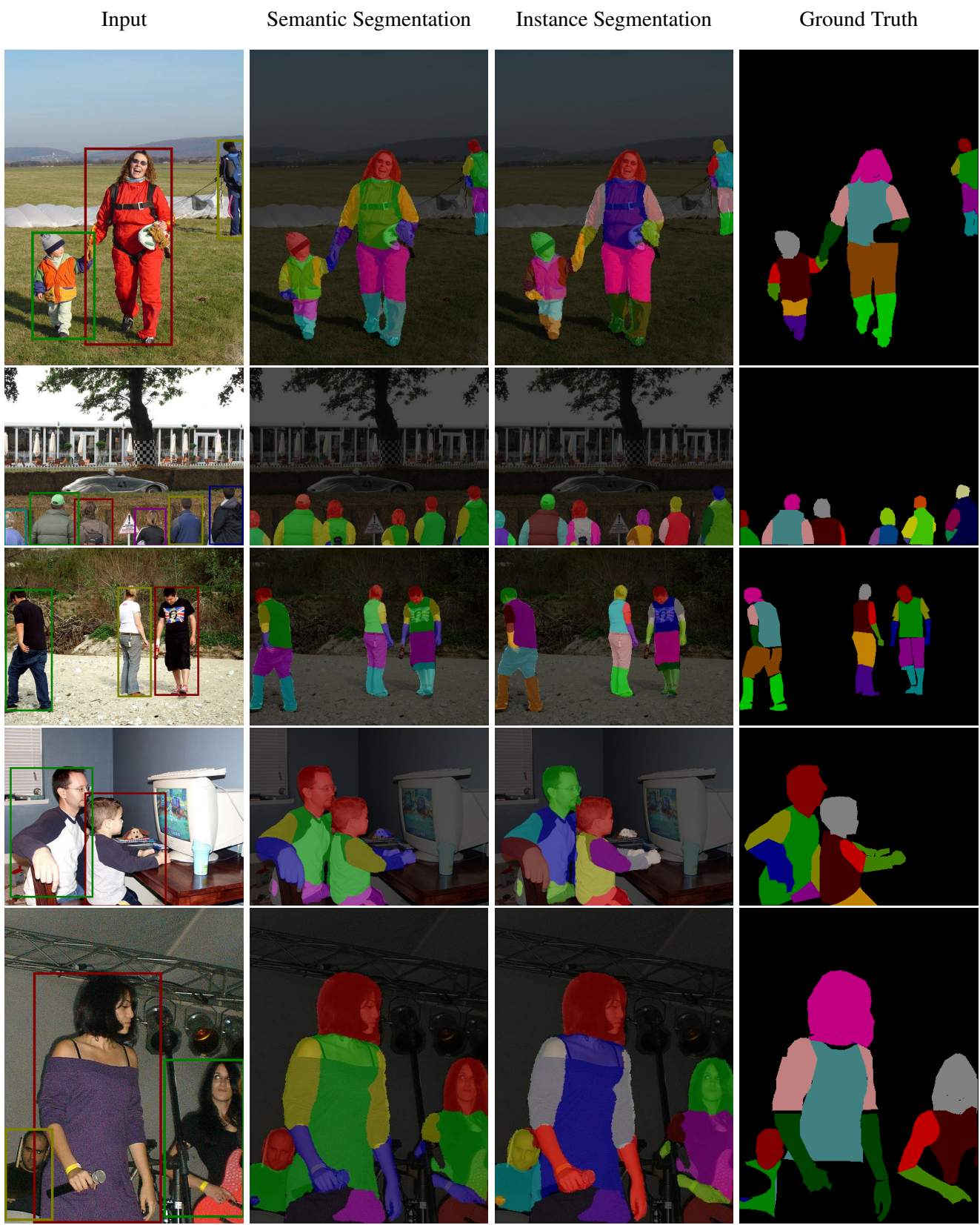

Figure 6: **Success cases of our method.** The first column shows the input image and the input detections we obtained from training the R-FCN detector [11]. The second column shows our final semantic segmentation (as described in Sec. 3.4 of the main paper). Our proposed method is able to leverage an initial category-level segmentation network and human detections to produce accurate instance-level part segmentation as shown in the third column.

Input          Semantic Segmentation          Instance Segmentation          Ground Truth

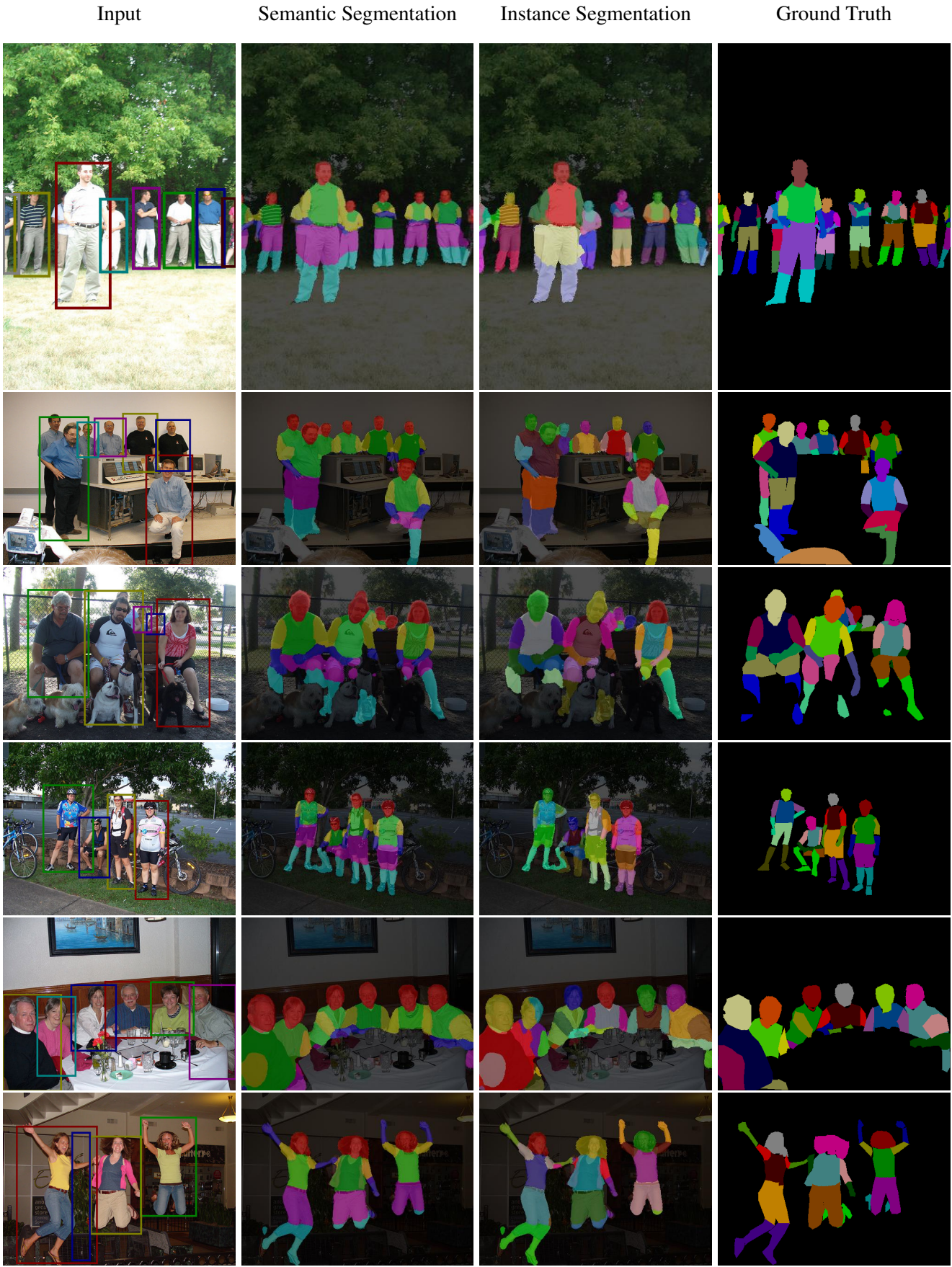

Figure 7: **Failure cases of our method.** *First three rows*: a missing human detection confuses the instance-level segmentation module. *Fourth and fifth row*: overlapping detection bounding boxes lead to incorrect instance label assignment when the overlapping region are visually similar. *Sixth row*: although our method is robust against false positive detections, two small regions on the leftmost person's left arm and left knee are assigned to the false positive detection.

Input MNC [10] Ours Ground Truth

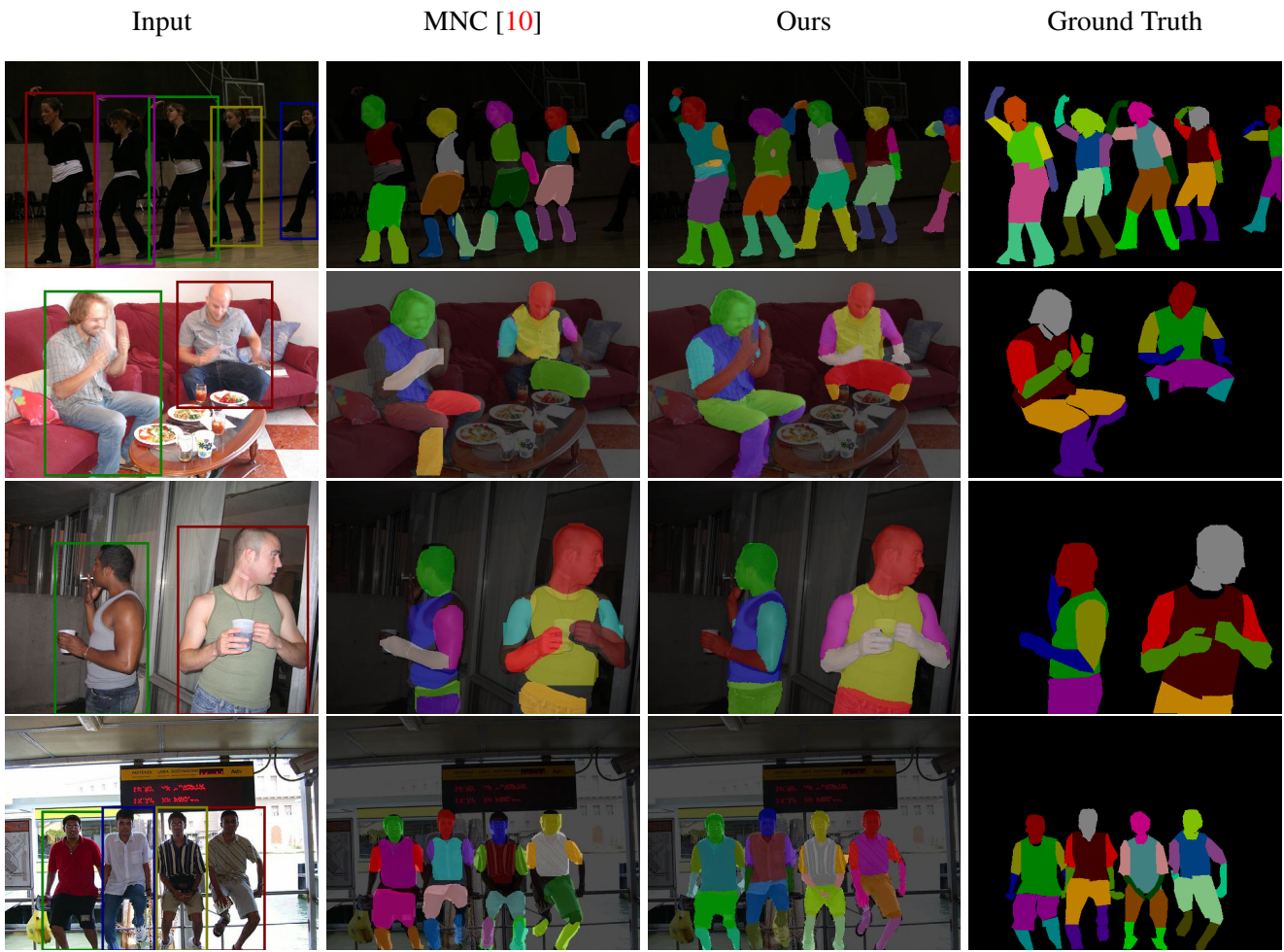

Figure 8: **Comparison to MNC on the Pascal Person-Parts [8] test set.** *First row*: unlike MNC which predicts for each part instance independently, our method reasons globally and jointly. As a result, MNC predicts two instances of lower legs for the same lower leg of the second and third person from the left. Furthermore, with a dedicated category-level segmentation module, we are less prone to false negatives, whereas MNC misses the legs of the rightmost person, and the lower arm of the second person from the right. *Second row*: while we can handle poor bounding box localisation because of our global potential term, MNC is unable to segment regions outside the bounding boxes it generates. Consequently, only one lower arm of the person on the left is segmented as the other one is outside the bounding box. The square corners of the segmented lower arm correspond to the limits imposed by the bounding box which MNC internally uses (box generation is the first stage of the cascade [10]). *Third row*: By analysing an image globally and employing a differentiable CRF, our method can produce more precise boundaries. As MNC does not perform category-level segmentation over the entire image, it has no incentive to produce a coherent and continuous prediction. Visually, this is reflected in the gaps of "background" between body parts of the same person. *Fourth row*: MNC predicts two instances of lower leg for the second person from the right, and fails to segment any lower arms for all four people due to the aforementioned problems.

| Input | MNC [10] | Ours | Ground Truth |
|---|---|---|---|

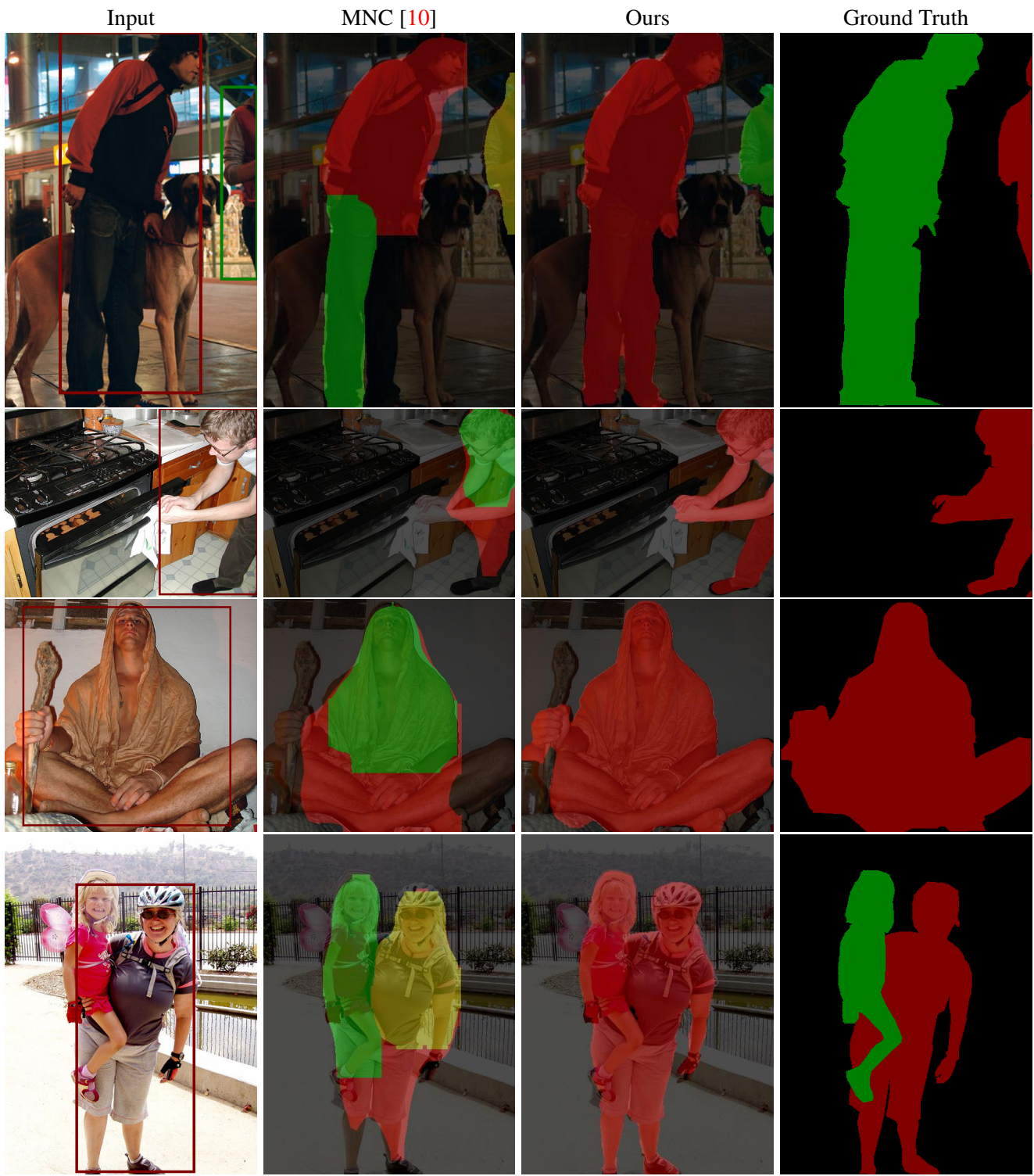

Figure 9: **Comparison to MNC on the Pascal Person-Parts [8] test set for instance-level human segmentation.** To generate the results in the second column, we run the public MNC model trained on VOC 2011/SBD [19] using the default parameters and extract only its human instance predictions. In contrast with proposal-driven methods such as MNC, our approach assigns each pixel to only one instance, is robust against non-ideal bounding boxes, and often produces better boundaries due to the Instance CRF which is trained end-to-end. *First and second row*: since MNC predicts instances independently, it is prone to predicting multiple instances for a single person. *Third row*: due to the global potential term, we can segment regions outside of a detection bounding box which fails to cover the entire person, whereas MNC is unable to recover from such imperfect bounding boxes, leading to its frequent occurrences of truncated instance predictions. *Fourth row*: a case where MNC and our method show different failure modes. MNC predicts three people where there are only two, and our method can only predict one instance due to a missing detection.

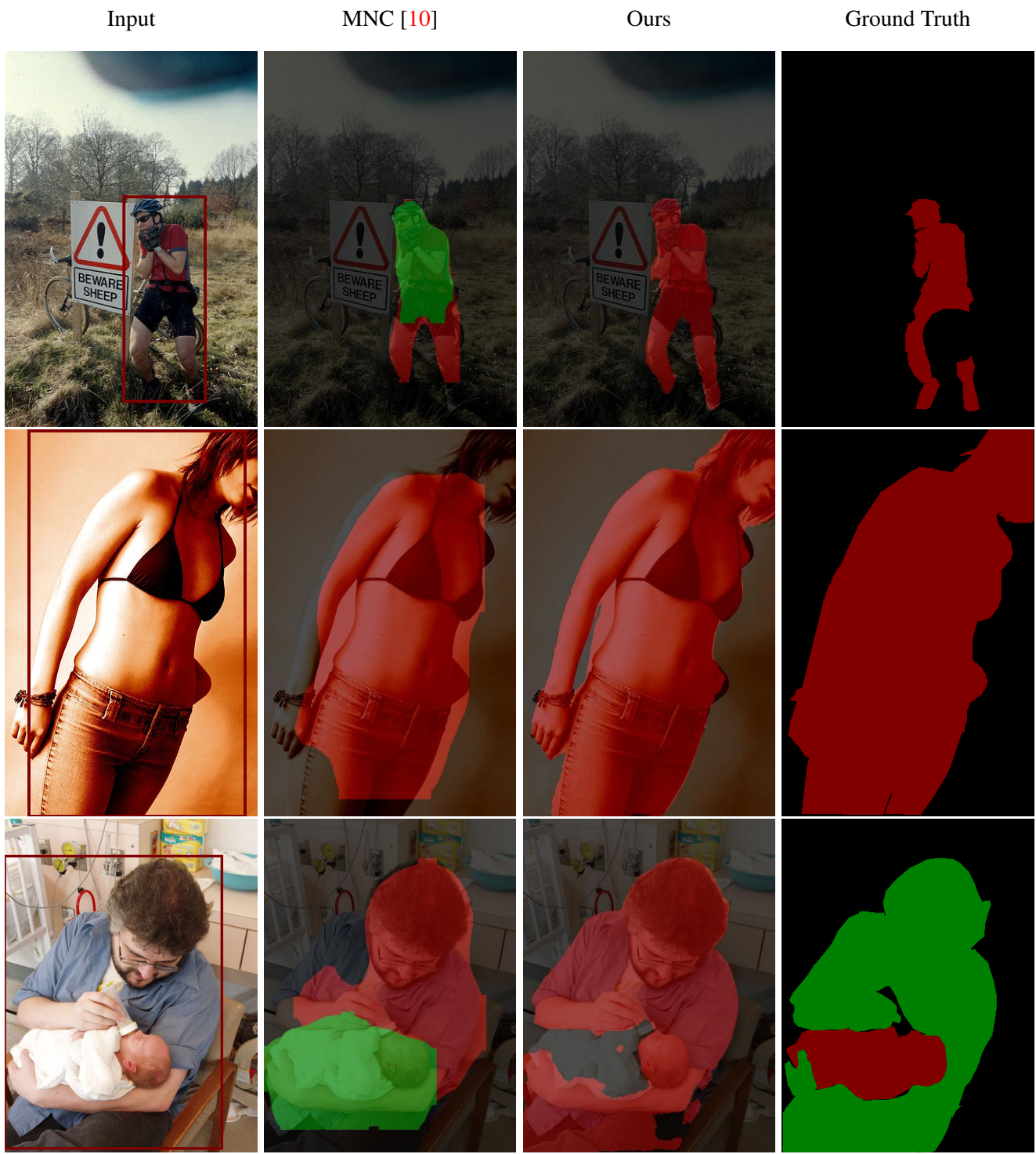

Figure 9 (Continued): **Comparison to MNC on the Pascal Person-Parts [8] test set for instance-level human segmentation.** *First row*: MNC is unable to recover from a false positive detection and predicts two people. *Second row*: while both MNC and our method start off with poor bounding box localisation that does not cover the whole instance, we are able to segment the entire person, whereas MNC is bounded by its flawed region proposal. *Third row*: MNC performs better in this case as it is able to segment the infant, whereas we miss her completely due to a false negative person detection.

# B    Additional information

We detail our initial category-level segmentation module and compare it to DeepLab-v2 [6] in Sec. B.1, present our network training details in Sec. B.2, and finally describe how we train the MNC model which serves as our baseline in Sec. B.3.

## B.1    Details of the category-level segmentation module

As shown in Fig 10b, the structure of our category-level segmentation module consists of a ResNet-101 backbone, and a classifier that extracts multi-scale features from the ResNet-101 output by using average pooling with different kernel sizes. While our category-level segmentation module and the Deeplab-v2 network (Fig. 10a) of Chen *et al.* [6] both attempt to exploit multi-scale information in the image, the approach of [6] entails executing three forward passes for each image, whereas we only need a single forward pass.

In comparison to Deeplab-v2, our network saves both memory and time, and achieves better performance. To carry out a single forward pass, our network uses 4.3GB of memory while Deeplab-v2 [6] needs 9.5GB, 120% more than ours. Speed-wise, our network runs forward passes at 0.255 seconds per image (3.9 fps), whereas Deeplab-v2 takes 55% longer, at 0.396 seconds per image (2.5 fps) on average. When Deeplab-v2 adds a CRF with 10 mean-field iterations to post-process the network output, it gains a small improvement in mean IoU by 0.54% [6], but it requires 11.2GB of memory to make a forward pass (140% of the total amount used by our full network including the instance-level segmentation module), and takes 0.960 seconds per image (1.0 fps), almost a quater of our frame rate. Tests are done on a single GeForce GTX Titan X (Maxwell) card. Overall, we are able to achieve better segmentation accuracy (as shown in Tab. 3 of our main paper) and is more memory- and time-efficient than Deeplab-v2.

## B.2    Training our proposed network

### B.2.1    Training the category-level segmentation module

We initialise our semantic segmentation network with the COCO pre-trained ResNet-101 weights provided by [6]. Training is first performed on the Pascal VOC 2012 training set using the extra annotations from [19], which combine to a total of 9012 training images. Care is taken to ensure that all images from the Pascal Person-Parts test set is excluded from this training set. A polynomial learning rate policy is adopted such that the effective learning rate at iteration $i$ is given by $l_i = l_0(1 - \frac{i}{i_{max}})^p$, where the base learning rate, $l_0$, is set to $6.25 \times 10^{-4}$, the total number of iterations, $i_{max}$, is set to 30k, and the power, $p$, is set to 0.9. A batch size of 16 is used. However, due to memory constraints, we simulate this batch size by "accumulating gradients": We carry out 16 forward and backward passes with one image per iteration, and only perform the weight update after completing all 16 passes. We use a momentum of 0.9 and weight decay of $1 \times 10^{-4}$ for these experiments. After 30k of iterations are completed, we take the best performing model and finetune on the Pascal Person-Parts training set using the same training scheme as described above. Note that the parameters of the batch normalisation modules are kept unchanged in the whole learning process.

Online data-augmentation is performed during training to regularise the model. The training images are randomly mirrored, scaled by a ratio between 0.5 and 2, rotated by an angle between -10 and 10 degrees, translated by a random amount in the HSV colour space, and blurred with a randomly-sized Gaussian kernel, all on-the-fly. We observe that these techniques are effective at reducing the accuracy gap between training and testing, leading to overall higher test accuracies.

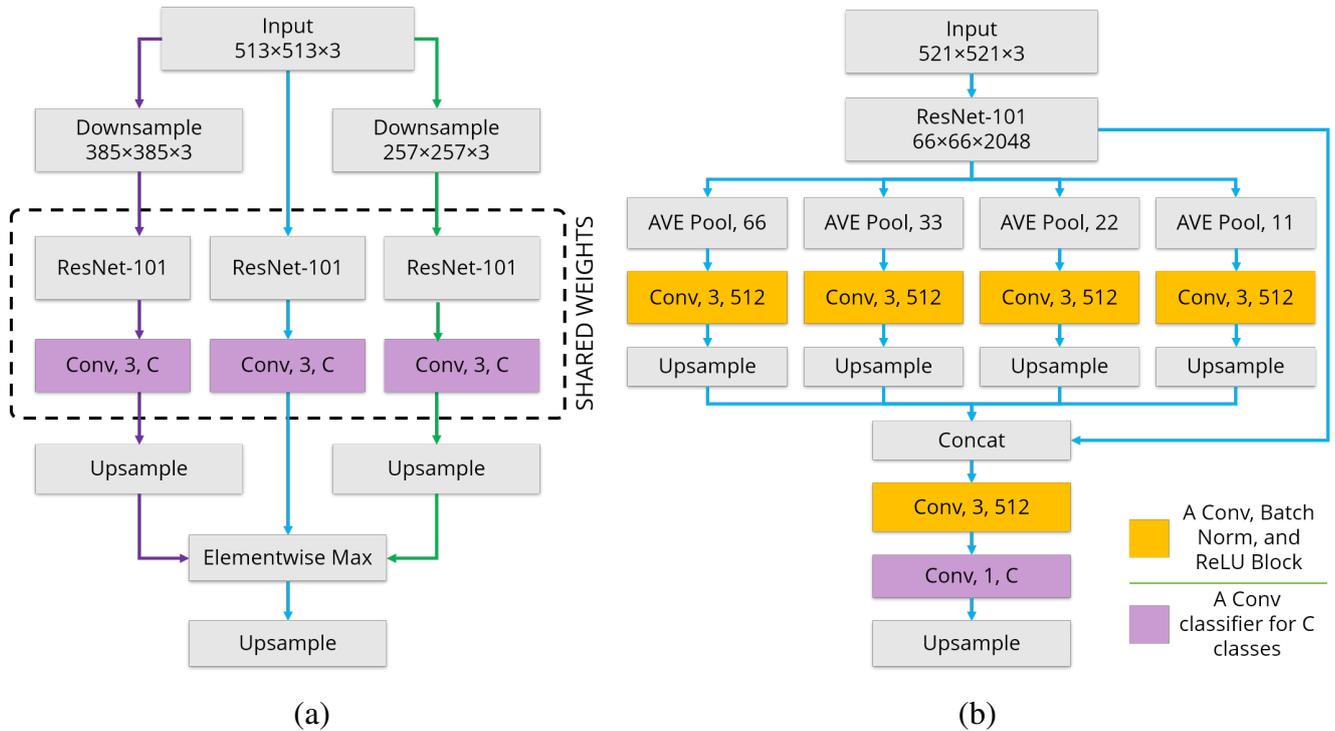

(a)                    (b)

Figure 10: Comparison of the Deeplab-v2 network structure which achieves 64.9% IoU on the Pascal Person-Parts dataset [6] and our network structure. The numbers following the layer type denote the kernel size and number of filters. For pooling layers, only their kernel sizes are shown as the number of filters is not applicable. The upsampling ratios can be inferred from the context. Fig. 10a: in the Deeplab-v2 architecture, a 513×513×3 input image is downsampled by two different ratios (0.75 and 0.5) to produce multi-scale input at three different resolutions. The three resolutions are independently processed by a ResNet-101-based network using shared weights (shown by the individually coloured paths). The output feature maps are then upsampled where appropriate, combined by taking the elementwise maximum, and finally upsampled back to 513×513. Fig. 10b: the category-level segmentation module proposed in this paper forwards an input image of size 521×521×3 through a ResNet-101-based CNN, producing a feature map of resolution 66×66×2048. This feature map is average-pooled with four different kernel sizes, giving us four feature maps with spatial resolutions 1×1, 2×2, 3×3, and 6×6 respectively. Each feature map undergoes convolution and upsampling, before being concatenated together with each other and the 66×66×2048 ResNet-101 output. This is followed by a convolution layer that reduces the dimension of the concatenated features to 512, and a convolutional classifier that maps the 512 channels to the size of label space in the dataset. Finally, the prediction is upsampled back to 521×521. In both Fig. 10a and 10b, the ResNet-101 backbone uses dilated convolution such that its output at `res5c` is at 1/8 of the input resolution, instead of 1/32 for the original ResNet-101 [22]. The convolutional classifiers (coloured in purple) output $C$ channels, corresponding to the number of classes in the dataset including a background class. For the Pascal Person-Parts Dataset, $C$ is 7. Best viewed in colour.

### B.2.2   Training the instance-level segmentation module

In our model, the pairwise term of the fully-connected CRF takes the following form:

$$\psi_{Pairwise}(v_i, v_j) = \mu(v_i, v_j)k(\mathbf{f_i}, \mathbf{f_j}) \qquad (5)$$

where $\mu(\cdot, \cdot)$ is a compatibility function, $k(\cdot, \cdot)$ is a kernel function, and $\mathbf{f_i}$ is a feature vector at spatial location $i$ containing the 3-dimensional colour vector $I_i$ and the 2-dimensional position vector $p_i$ [24].

We further define the kernel as follows:

$$k(\mathbf{f}_i, \mathbf{f}_j) = w^{(1)}\exp\Big(-\frac{|p_i - p_j|^2}{2\theta_\alpha^2} - \frac{|I_i - I_j|^2}{2\theta_\beta^2}\Big) + w^{(2)}\exp\Big(-\frac{|p_i - p_j|^2}{2\theta_\gamma^2}\Big) \qquad (6)$$

where $w^{(1)}$ and $w^{(2)}$ are the linear combination weights for the bilateral term and the Gaussian term respectively. In order to determine the initial values for the parameters in the Instance CRF to train from, we carry out a random search. According to the search results, the best prediction accuracy is obtained by initialising $w^{(1)} = 8$, $w^{(2)} = 2$, $\theta_\alpha = 2$, $\theta_\beta = 8$, $\theta_\gamma = 2$. Furthermore, we use a fixed learning rate of $1 \times 10^{-6}$, momentum of 0.9, and weight decay of $1 \times 10^{-4}$ for training both the instance-level and category-level segmentation modules jointly. Although we previously use the polynomial learning rate policy, we find that for training the instance-level segmentation module, a fixed learning rate leads to better results. Furthermore, our experiments show that a batch size of one works best at this training stage. Using this scheme, we train for 175k iterations, or approximately 100 epochs.

## B.3 Training Multi-task Network Cascades (MNC)

We use the publicly available Multi-task Network Cascades (MNC) framework [10], and train a new model for instance-level part segmentation using the Pascal Person-Parts dataset. The weights are initialised with the officially released MNC model[1] which has been trained on Pascal VOC 2011/SBD [19]. The base learning rate is set to $1 \times 10^{-3}$, which is reduced by 10 times after 20k iterations. A total of 25k training iterations are carried out. A batch size of 8, momentum of 0.9 and weight decay of $5 \times 10^{-4}$ are used. These settings are identical to the ones used in training the original MNC and provided in their public source code. Using these settings, we are also able to reproduce the experimental results obtained in the original MNC paper [10], and hence we believe that the MNC model we have trained acts as a strong baseline for our proposed approach.

---

[1] https://github.com/daijifeng001/MNC